\documentclass{article}
\usepackage{spconf,amsmath,amssymb,graphicx}
\usepackage{hyperref}

\title{KinshipGAN: Synthesizing of Kinship Faces From Family Photos by Regularizing a Deep Face Network}
\name{Savas Ozkan$^1$, Akin Ozkan$^2$}
\address{1. Middle East Technical University, Department of Electrical/Electronics Engineering, Ankara, Turkey\\
	     2. Atilim University, Department of Electrical/Electronics Engineering, Ankara, Turkey\\}
\begin{document}
%\ninept
%
\maketitle
\begin{abstract}
In this paper, we propose a kinship generator network that can synthesize a possible child face by analyzing his/her parent's photo. For this purpose, we focus on to handle the scarcity of kinship datasets throughout the paper by proposing novel solutions in particular. To extract robust features, we integrate a pre-trained face model to the kinship face generator.  Moreover, the generator network is regularized with an additional face dataset and adversarial loss to decrease the overfitting of the limited samples. Lastly, we adapt cycle-domain transformation to attain a more stable results. Experiments are conducted on Families in the Wild (FIW) dataset. The experimental results show that the contributions presented in the paper provide important performance improvements compared to the baseline architecture and our proposed method yields promising perceptual results.
\end{abstract}
\section{Introduction}
\label{sec:intro}
\vspace{-0.2cm}

A human brain can verify kinship from photos by analyzing the disriminative patterns of facial parts. This feature is a strong evidence that how brain fascinatingly complex it is. Recently, an immense number of methods have been proposed to achieve kinship verification by computers, since learning-based deep models have shown impressive powers to extract these latent patterns automatically from faces~\cite{vggface_2015, kin1_2017, kin2_2017}. In particular, these methods  outperform the performance achieved by humans for various identification problem~\cite{alexnet_2012, vggface_2015}. Ultimately, the outputs of the models can be used for the identification of missing people, child/parent search as well as tracking some statistics for recommendation services. 

However, looking the problem in reverse, more intuitively, guessing possible child faces by analyzing their parent photos, is not quite motivated as the original problem  in literature (i.e., recognition and verification). To the best of our knowledge, there is also a limited interest to tackle the problem~\cite{syn_2017}, even if there are several promising methods to synthesize human faces from large-data collections based on generative deep models~\cite{gan_2014, dcgan_2015, began_2017}.

In general, the objective of this problem (i.e. ,synthesizing kinship face) is that for the given input of a parent photo (either mother or father), a method synthesizes the most probable faces of a child by exploiting latent facial features exhibited on the parents' faces. However, the robustness of the models, especially for deep models, strongly depends on the number of training samples and the diversity of the datasets. Moreover, currently available datasets for kinship verification are quite small and models should be regularized based on this limitation so as to achieve perceptually satisfying results.

In this paper, we propose a fully convolutional network (FCN) which transforms a parent face in a latent space with the responses of encoder layers and iteratively decodes these responses to reconstruct a possible kinship face. For this purpose, we present three novel contributions to the standard FCN for kinship face synthesis: 1) We use a pre-trained network for the encoder layers which is optimized for face recognition on a large-scale dataset. Eventually, this allows us to extract more robust hidden features even if limited numbers of faces are modeled for face synthesis. 2)  Although use of the encoder layers provides several advantages such as sparsity for person identification,  decoder layers can easily overfit to the training data due to the large dimensionality of the hidden features. At the end, it hardens the problem to generalize an optimum solution for diverse face scenarios. Hence, we leverage adversarial loss with large-scale unsupervised data to mitigate the overfitting with its generalization capability. 3) Lastly, we employ cycle-domain transformation~\cite{cycle_2017} (i.e., transforming from parent-to-child as well as child-to-parent) which leads to more stable results.

The paper is organized as follows. First, we review the literature on face synthesis and kinship verification, since these steps are two major basis of our problem. Later, the details of the proposed method are presented for kinship synthesis. Lastly, experimental results are reported and we explain the final remarks of the paper.

\section{Related Work}
\label{sec:format}

In this section, face synthesis and kinship verification are reviewed in detail, since these are two critical ingredients for an effective kinship synthesis. 

\noindent
\textbf{Face Synthesis}: The early studies on face synthesis are initially presented for hallucinating faces from low-resolution images to infer their high-frequency details~\cite{hal_2000, hal_2001}. In these works, common characteristics of faces such as eyes, mount and symmetry are particularly enhanced. However, their main limitation is that the solution strictly relies on data (i.e no generalization capacity) and natural image manifold learning (i.e., memorizing) can be stuck to the case that it only transforms image patches from low-resolutions to higher ones by taking averages of all possible solutions at the end. Similarly, autoencoder-based (AE) methods have the similar drawbacks for the solution. ~\cite{syn_2017} aims to generate kinship faces by promoting facial dynamics (i.e., expression) along with visual appearances based on AE, thus it is able to transfer personal expressions to prospective children.

Variational autoencoder (VAE)~\cite{vae_2013} is a probabilistic way of synthesizing images by computing random latent variables according to the input at the encoder layers. Thus, this practically improves the generalization of the models and attains diverse results for various image synthesis problems as well as faces. However, it still lacks to reach the complexity of the problems  (i.e., it underestimates the problem with the fixed sized parameters, i.e., mean and variance values). At the end, overly-smoothed results are obtained.

Recently, generative adversarial networks (GAN)~\cite{gan_2014, dcgan_2015, began_2017} yield perceptually impressive results for image generation. In particular, face synthesis can be achieved in an unsupervised manner by incorporating various poses, expressions, genders, skin colors, and hair types. Moreover, it allows users to transform images to different domains by simply conditioning the solution~\cite{i2i_2017, star_2017}. The superiority of GAN over VAE/AE is explained in~\cite{adv_2016} which remarks that GAN preserves the fine-detail solution about the problem, while VAE/AE approximates it roughly. 

\noindent
\textbf{Kinship Verification}: Kinship verification/recognition is initially based hand-crafted shallow facial features by incorporating skin color and/or higher-order gradient patterns exhibited from facial photos~\cite{verif1_2014, verif2_2014}. Moreover, use of videos instead of single images is explored~\cite{vid_2013} and the authors assert that it can be useful to verify faces with spatio-temporal appearances, implicitly facial expressions.

Recently, deep models attain state-of-the-art performance for the problem~\cite{ kin3_2015, kin1_2017, kin2_2017, kin3_2017}. In general, their solutions are based on transferring the trainable parameters from an available face model and finetuning with kinship data due to the scarcity of samples. Lastly, feature space is frequently learned with a triplet loss similar to face recognition problem~\cite{vggface_2015}.

\section{Kinship Face Synthesis}
\label{sec:pagestyle}

For a given parent photo $x \in \mathbb{R}^{128 \times 128 \times 3}$ (i.e., after face detector\footnote{Dlib library is used for face detection. \url{http://dlib.net/}}), our objective is to synthesis a child face $y \in \mathbb{R}^{128 \times 128 \times 3}$ by exploiting the responses of a Generator network $G_c(.)$ that consists of fully convolutional encoder and decoder layers. To flexibly define different genders for prospective child faces (i.e male or female), $y$ is conditioned at the decoder layer by a label $c$ formulated as $\hat{y} = G_c(x, c)$. This condition is also boosted by an auxiliary classifier as in~\cite{star_2017} for more stable results. 

\begin{figure}[t]
\centering
\includegraphics[scale=0.3]{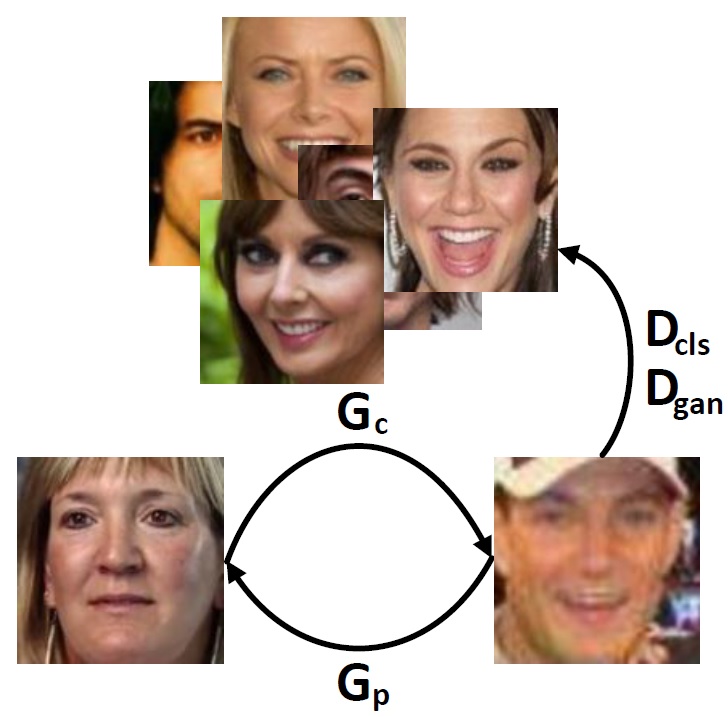}
\vspace{-0.2cm}
\caption{The flow of the proposed method.}
\vspace{-0.6cm}
\label{fig:flow}
\end{figure}

\noindent
\textbf{Generator Network}: Generator network $G$ consists of fully convolution layers and it is an AE architecture. It aims to extract representative latent abstracts from an input image and to generate a face based on these latent abstracts. To increase to information flow from the encoder to decoder layers, we employ skip-connections as in~\cite{res_2016}. Practically, these connections improve the perceptual quality and stability of the generated faces. 

Furthermore, we follow the similar observations about the architecture that are presented in~\cite{dcgan_2015} for generative network (i.e., normalization layer with leaky-ReLU activation). 

In addition, the target face is conditioned by label $c$ at the decoder layers which enables us to define the gender of the generated kinship face based on this label as either male or female. However, for better perceptual distinction, we add a penalty term to the loss function which will be explained in the following section in detail.

\noindent
\textbf{Loss Functions}: 
For parameter optimization, first, we adapt the assumption of matching high-level activations of a pre-trained network between the original and generated faces instead of pixel differences~\cite{style_2016, per_2016} (i.e., renowned as content loss). Ultimately, this loss allows us to preserve latent content which can be exhibited from faces (i.e., it enables to transfer high-level distinct facial features from larger receptive fields than individual pixels). In addition, it provides robustness to some cases which contain not perfectly registered faces and/or severe pose variations.  You can find further discussions about the assumption in~\cite{refi_2017}.

The layer activations $\phi_l(.) (l =4,5)$ of VGGFace-16~\cite{vggface_2015} (i.e., conv4\_3 and conv5\_3) are used and the loss between a training pair ($x, y$) is minimized as:
\begin{eqnarray}
\label{eqn:lossc}
\mathcal{L}_{con}^C =  \sum_{l} || \phi_l(y) - \phi_l(  G_c(x, c) ) ||_1.
\end{eqnarray}

Note that by incorporating only higher layer activations in the loss function, we opt to preserve global facial part similarities than fine-details by which synthesis of fine-details is quite difficult by just analyzing images as expected (i.e., some peripheral dependencies).

Moreover, we introduce an auxiliary classifier to condition the gender of faces. This network $D_{cls}$ categories an input face based on a softmax cross-entropy loss as male or female and it propagates the error to the network for Generative and Discriminative layers:
\begin{eqnarray}
\label{eqn:lossc}
\mathcal{L}_{aux}^D = - \log (D_{cls}(c|y)).
\end{eqnarray}
\begin{eqnarray}
\label{eqn:lossc}
\mathcal{L}_{aux}^G = - \log (D_{cls}(y|c)).
\end{eqnarray}

\noindent
\textbf{Face Encoder Layers}: As mentioned, the main drawback of synthesizing a kinship face is that there is a limited number of training samples compared to the other facial problems/datasets in literature~\cite{kin1_2017, kin2_2017}. Moreover, learning methods need sufficiently large and diverse samples to obtain reliable models. Eventually, these issues weaken the generalization power of the methods by overfitting to the limited samples. 

Therefore, instead of creating a model from scratch for encoder layer, we replace the layer parameters of the encoder network with a pre-trained model (i.e., VGGFace-16). Note that this model is learned on a large and diverse face dataset~\cite{vggface_2015}. Practically, it enables to extract more discriminative and latent features about faces. Also, it adds robustness against noise and pose variations. Lastly, these parameters are not finetuned during the learning stage.

Remark that use of a pre-trained model has another advantage that the generated kinship faces can automatically transfer the facial expressions of parents from photos. Thus, there is no need to utilize a different loss function or conditional labels for the network (Please see the experimental results).

\noindent
\textbf{Adversarial Loss}: The Generator network computes large-dimensional hidden abstracts about data. Furthermore, the total number of operations is multiplied when skip-connections are utilized. However, even if rich representations are extracted, the sparsity of the abstracts disturbs the stability and convergence of the parameters to an optimum solution for small image sets. ~\cite{agg_2015} explain that mapping large dimensionality to a lower space, in other words, willingly degrading the sparsity with a reduction method (i.e., PCA etc.), can definitely improve the performance of deep convolution methods for various transfer learning problems.

Based on this observation, we improve the generalization capacity and stability of the Generator network with adversarial network scheme trained on a different and larger face dataset. Thus, GAN replaces the reduction method and it acts like a degradation function to obtain indistinguishable faces. 

For this purpose, we employ energy-based Wasserstein GAN objective~\cite{began_2017} on a sample $x_{cp}$ drawn from a different face set (i.e., CelebA dataset~\cite{celeb_2015}) which is formally defined as:
\begin{eqnarray}
\label{eqn:lossc}
\mathcal{L}^D_{gan} = D_{gan}(x_{cp})- k_t. D_{gan}(G_c(x_{cp}, c)),
\end{eqnarray}
\vspace{-0.5cm}
\begin{eqnarray}
\label{eqn:lossc}
\mathcal{L}^G_{gan} = D_{gan}(G_c(x_{cp}, c)).
\end{eqnarray}

\noindent
where super-scripts of the losses indicate which parameter sets are updated (Generator (G) or Discriminator (D)). Moreover, $D_{gan}$ (i.e., discriminator) is structured as an autoencoder network and the reconstruction-based loss is utilized. Lastly, $k_t$ is a trainable parameter and $\gamma$ of~\cite{began_2017} is set to 0.7 to diversify the generated faces. Note that the network is updated by considering content loss and generative loss simultaneously at the end.

\noindent
\textbf{Cycle-Consistency}: Lastly, we employ a cycle-domain transformation as in~\cite{cycle_2017} to achieve more stable results. By this way, a consistent facial transformation can be obtained by linking the generated kinship face by his/her parent face. For parameter optimization, we add an additional cost term similar to Eq. 1: 
\begin{eqnarray}
\label{eqn:lossc}
\mathcal{L}_{con}^P =  \sum_{l} || \phi_l(x) - \phi_l(  G_p(G_c(x, c) )) ||_1.
\end{eqnarray}

\noindent
$G_p(.)$ has shared encoder parameters with $G_c(.)$ while different decoder parameters.

\noindent
\textbf{Full Method}: 
Finally, the overall flow of the proposed method is illustrated in Fig.~\ref{fig:flow}. Furthermore, the full objective functions for Generative $\mathcal{L}_G$ and Discriminative $\mathcal{L}_D$ layers are written respectively as:
\begin{eqnarray}
\label{eqn:lossc}
\mathcal{L}_{D} = \lambda_{gan} \mathcal{L}^D_{gan} + \lambda_{aux} \mathcal{L}_{aux}^D,
\end{eqnarray}
\vspace{-0.5cm}
\begin{eqnarray}
\label{eqn:lossc}
\mathcal{L}_{G} = \lambda_{c} \mathcal{L}_{con}^C + \lambda_{p} \mathcal{L}_{con}^P + \mathcal{L}^G_{gan} +  \lambda_{aux} \mathcal{L}_{aux}^G.
\end{eqnarray}

\noindent
where $\lambda_{gan}$, $\lambda_{c}$, $\lambda_{p}$ and $\lambda_{aux}$ control the influences of the loss functions. Empirically, they are set to 10, 0.1, 0.001 and 0.1 by taking the parameter overfitting into account for small datasets. 

\section{Experiments}
\label{sec:typestyle}

\begin{figure*}[t]
\vspace{-0.7cm}
\centering
\includegraphics[scale=0.4]{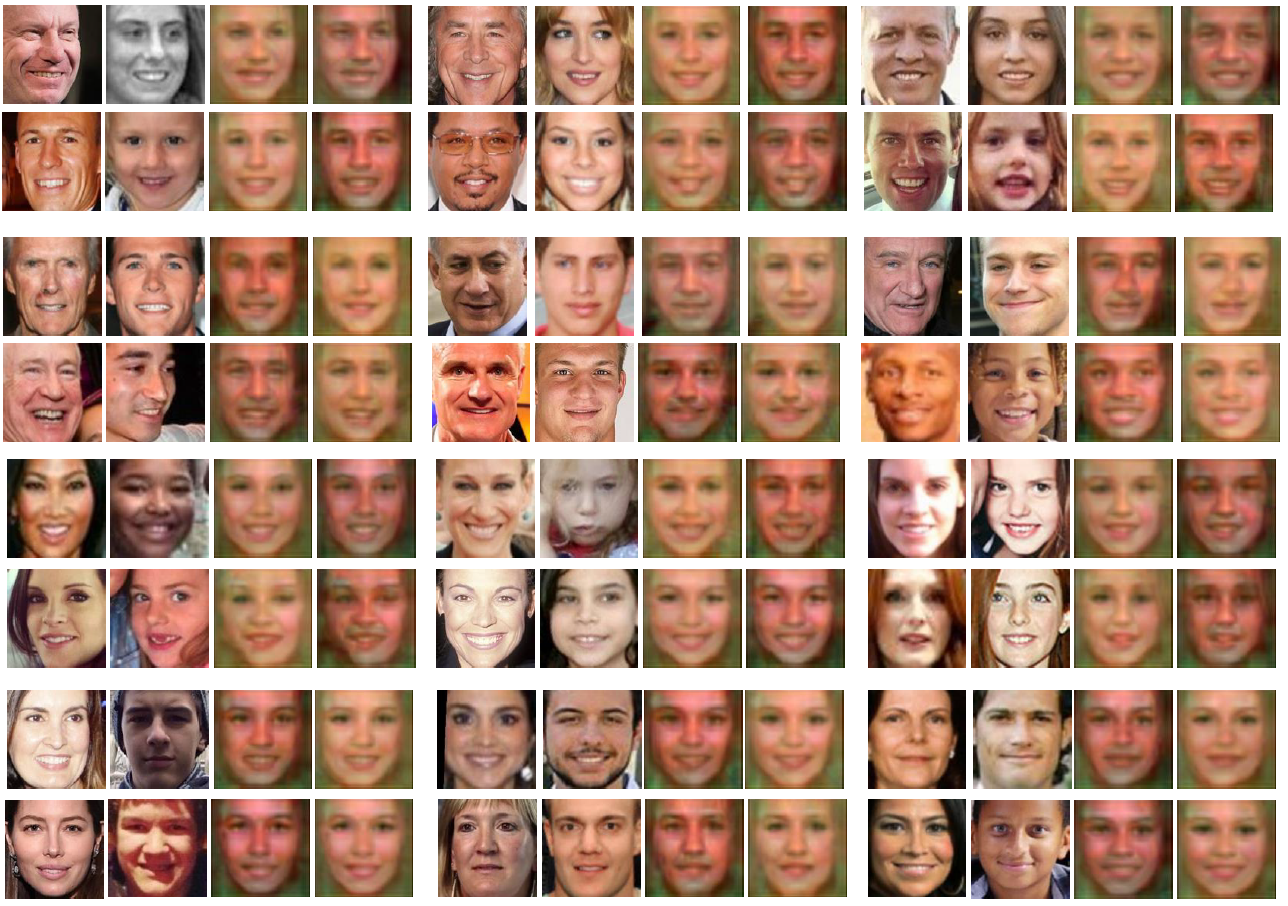}
\vspace{-0.2cm}
\caption{Several visual outputs of the proposed method for father-daughter, father-son, mother-daughter and mother-son. You can also find the opposite gender of the results.}
\vspace{-0.4cm}
\label{fig:test}
\end{figure*}

\noindent
\textbf{Datasets}: For kinship synthesis, we use the dataset released for Large-Scale Kinship Recognition Data Challenge and it is called as Families in the Wild (FIW)~\cite{robinson2018visual}. This dataset comprises approximately 600K face pairs from 300 families in the training set. Since the labels of test set are not available, we use the validation set (randomly select 20K) to evaluate the performance of the proposed method. Moreover, we use only father-son, father-daughter, mother-son and mother-daughter relations. In addition, we utilize CelebA dataset~\cite{celeb_2015} to regularize the network which contains 200K celebrity images with 40 different attributes. Note that we exploit only gender attribute of faces in this paper (i.e., male or female).

\begin{table}[b]
\begin{center}
\vspace{-0.2cm}
\begin{tabular}{ l c }
\hline \hline
Model &  Top-100 Accuracy \\ 
\hline 
KinshipGAN (w$\setminus$o Deep Face)      &  0.048 \\ 
KinshipGAN ($\lambda_c=1.0$)                &  0.063   \\ 
KinshipGAN ($\lambda_c=0.1$)    & \textbf{0.107}  \\ 
\hline \hline
\end{tabular}

\vspace{-0.2cm}
\caption{Top-100 retrieval accuracy of kinship faces generated by the proposed method in FIW validation dataset.}
\vspace{-0.5cm}
\label{tab:test}
\end{center}
\end{table}

To evaluate the performance, we use NN search accuracy along with the qualitative results which calculates the total number of correct face ids at the top-100 of ranked lists for given synthesis faces to the system. Additionally, fc6 layer activations of VGGFace-16 are used to represent the faces and cosine distance is computed.

\noindent
\textbf{Implementation Details}: The parameters are optimized by Adam optimizer~\cite{adam_2014} and learning rate is fixed to 0.0001. Moreover, batch size and iteration numbers are set to 32 and 100K respectively. All codes are implemented on TensorFlow deep learning framework.

\subsection{Experimental Results}
\label{ssec:subhead}

Table~\ref{tab:test} shows the retrieval accuracy of kinship faces generated by our method on real validation face dataset. Even if the methods do not obtain significant performance for the retrieval (which is not main scope of the method), it can clearly illustrate the effects of the contributions presented in this paper to the kinship face synthesis. In particular, use of a pre-trained face model for the encoder part introduces a noticeable distinguishable power to the model. Furthermore, setting $\lambda_c$ coefficient to higher values (e.g., 1.0) which determines the influence of face similarities, can adversely affect the performance and perceptually worse results can be obtained.

Fig.~\ref{fig:test} illustrates the output of the proposed method for father-daughter, father-son, mother-daughter and mother-son relations (You can find additional results on Appendix). Furthermore, the opposite of the gender for each face is also given. From these results, the proposed method yields perceptually promising results on kinship face synthesis. Particularly, the method can preserve the facial expression, pose etc. exhibited from parent photos by the cycle-domain transformation and  the pre-trained network.

\section{Conclusion}
\label{sec:majhead}

In this paper, we propose a kinship face generator network that can yield promising results under the scarcity of kinship samples. Throughout the paper, we propose three main contributions. First, in order to extract robust facial features, we exploit a pre-trained deep face model in the network. Later, adversarial scheme is used to improve the generalization capacity of the network and to prevent the overfitting. Lastly, cycle-domain transformation approach is utilized to provide consistency between parent-to-child translation.  The experimental results show that the proposed method achieves promising perceptual results.

\bibliographystyle{IEEEtrans}
\bibliography{Template}

% Generated by IEEEtranS.bst, version: 1.14 (2015/08/26)
\begin{thebibliography}{10}
\providecommand{\url}[1]{#1}
\csname url@samestyle\endcsname
\providecommand{\newblock}{\relax}
\providecommand{\bibinfo}[2]{#2}
\providecommand{\BIBentrySTDinterwordspacing}{\spaceskip=0pt\relax}
\providecommand{\BIBentryALTinterwordstretchfactor}{4}
\providecommand{\BIBentryALTinterwordspacing}{\spaceskip=\fontdimen2\font plus
\BIBentryALTinterwordstretchfactor\fontdimen3\font minus
  \fontdimen4\font\relax}
\providecommand{\BIBforeignlanguage}[2]{{%
\expandafter\ifx\csname l@#1\endcsname\relax
\typeout{** WARNING: IEEEtranS.bst: No hyphenation pattern has been}%
\typeout{** loaded for the language `#1'. Using the pattern for}%
\typeout{** the default language instead.}%
\else
\language=\csname l@#1\endcsname
\fi
#2}}
\providecommand{\BIBdecl}{\relax}
\BIBdecl

\bibitem{agg_2015}
A.~Babenko and V.~Lempitsky, ``Aggregating local deep features for image
  retrieval.'' \emph{IEEE ICCV}, 2015.

\bibitem{hal_2000}
S.~Baker and T.~Kanade, ``Hallucinating faces.'' \emph{IEEE FG}, 2000.

\bibitem{began_2017}
D.~Berthelot, T.~Schumm, and L.~Metz, ``Began: Boundary equilibrium generative
  adversarial networks.'' \emph{arXiv preprint arXiv:1703.1071}, 2017.

\bibitem{refi_2017}
Q.~Chen and V.~Koltun, ``Photographic image synthesis with cascaded refinement
  networks.'' \emph{IEEE ICCV}, 2017.

\bibitem{star_2017}
Y.~Choi, M.~Choi, M.~Kim, J.-W. Ha, S.~Kim, and J.~Choo, ``Stargan: Unified
  generative adversarial networks for multi-domain image-to-image
  translation.'' \emph{arXiv preprint arXiv:1711.09020}, 2017.

\bibitem{vid_2013}
H.~Dibeklioglu, A.~A. Salah, and T.~Gevers, ``Like father, like son: Facial
  expression dynamics for kinship verification.'' 2013.

\bibitem{per_2016}
A.~Dosovitskiy and T.~Brox, ``Generating images with perceptual similarity
  metrics based on deep networks.'' \emph{NIPS}, 2016.

\bibitem{kin2_2017}
Q.~Duan and L.~Zhang, ``Kinship verification with deep convolutional neural
  networks.'' \emph{ACM MM Workshops}, 2017.

\bibitem{adv_2016}
V.~Dumoulin, I.~Belghazi, B.~Poole, O.~Mastropietro, A.~Lamb, M.~Arjovsky, and
  A.~Courville, ``Adversarially learned inference.'' \emph{arXiv preprint
  arXiv:1606.00704}, 2016.

\bibitem{syn_2017}
I.~O. Ertugrul and H.~Dibeklioglu, ``What will your future child look like?
  modeling and synthesis of hereditary patterns of facial dynamics.''
  \emph{IEEE FG}, 2017.

\bibitem{style_2016}
L.~A. Gatys, A.~S. Ecker, and M.~Bethge, ``Image style transfer using
  convolutional neural networks.'' \emph{IEEE CVPR}, 2016.

\bibitem{gan_2014}
I.~Goodfellow, J.~Pouget-Abadie, M.~Mirza, B.~Xu, D.~Warde-Farley, S.~Ozair,
  A.~Courville, and Y.~Bengio, ``Generative adversarial nets.'' \emph{NIPS},
  2014.

\bibitem{res_2016}
K.~He, X.~Zhang, S.~Ren, and J.~Sun, ``Deep residual learning for image
  recognition.'' 2016.

\bibitem{i2i_2017}
P.~Isola, J.-Y. Zhu, T.~Zhou, and A.~A. Efros, ``Image-to-image translation
  with conditional adversarial networks.'' \emph{IEEE ICCV}, 2017.

\bibitem{adam_2014}
D.~Kingma and J.~Ba, ``Adam: A method for stochastic optimization.''
  \emph{arXiv preprint}, 2014.

\bibitem{vae_2013}
D.~P. Kingma and M.~Welling, ``Began: Boundary equilibrium generative
  adversarial networks.'' \emph{arXiv preprint arXiv:1312.6114}, 2013.

\bibitem{alexnet_2012}
A.~Krizhevsky, I.~Sutskever, and G.~E. Hinton, ``Imagenet classification with
  deep convolutional neural networks.'' 2012.

\bibitem{kin1_2017}
Y.~Li, J.~Zeng, J.~Zhang, A.~Dai, M.~Kan, S.~Shan, and X.~Chen, ``Kinnet:
  Fine-to-coarse deep metric learning for kinship verification.'' \emph{ACM MM
  Workshops}, 2017.

\bibitem{hal_2001}
C.~Liu, H.-Y. Shum, and C.-S. Zhang, ``A two-step approach to hallucinating
  faces: global parametric model and local nonparametric model.'' \emph{IEEE
  CVPR}, 2001.

\bibitem{celeb_2015}
Z.~Liu, P.~Luo, X.~Wang, and X.~Tang, ``Deep learning face attributes in the
  wild.'' \emph{IEEE ICCV}, 2015.

\bibitem{kin3_2017}
J.~Lu, J.~Hu, and Y.-P. Tan, ``Discriminative deep metric learning for face and
  kinship verification.'' 2017.

\bibitem{vggface_2015}
O.~M. Parkhi, A.~Vedaldi, and A.~Zisserman, ``Deep face recognition.'' 2015.

\bibitem{dcgan_2015}
A.~Radford, L.~Metz, and S.~Chintala, ``Unsupervised representation learning
  with deep convolutional generative adversarial networks.'' \emph{arXiv
  preprint arXiv:1511.06434}, 2015.

\bibitem{robinson2018visual}
J.~P. Robinson, M.~Shao, Y.~Wu, H.~Liu, T.~Gillis, and Y.~Fu, ``Visual kinship
  recognition of families in the wild,'' \emph{TPAMI}, 2018.

\bibitem{verif1_2014}
X.~Wang and C.~Kambhamettu, ``Leveraging appearance and geometry for kinship
  verification.'' 2014.

\bibitem{verif2_2014}
H.~Yan, J.~Lu, W.~Deng, and X.~Zhou, ``Discriminative multimetric learning for
  kinship verification.'' 2014.

\bibitem{kin3_2015}
K.~Zhang, Y.~Huang, C.~Song, H.~Wu, and L.~Wang, ``Kinship verification with
  deep convolutional neural networks.'' \emph{BMVC}, 2015.

\bibitem{cycle_2017}
J.-Y. Zhu, T.~Park, P.~Isola, and A.~A. Efros, ``Unpaired image-to-image
  translation using cycle-consistent adversarial networks.'' \emph{IEEE ICCV},
  2017.

\end{thebibliography}

\onecolumn

\section*{Appendix}
\subsection*{Additional Qualitative Results}

\begin{figure}[hbt]
\centering
\includegraphics[scale=0.75]{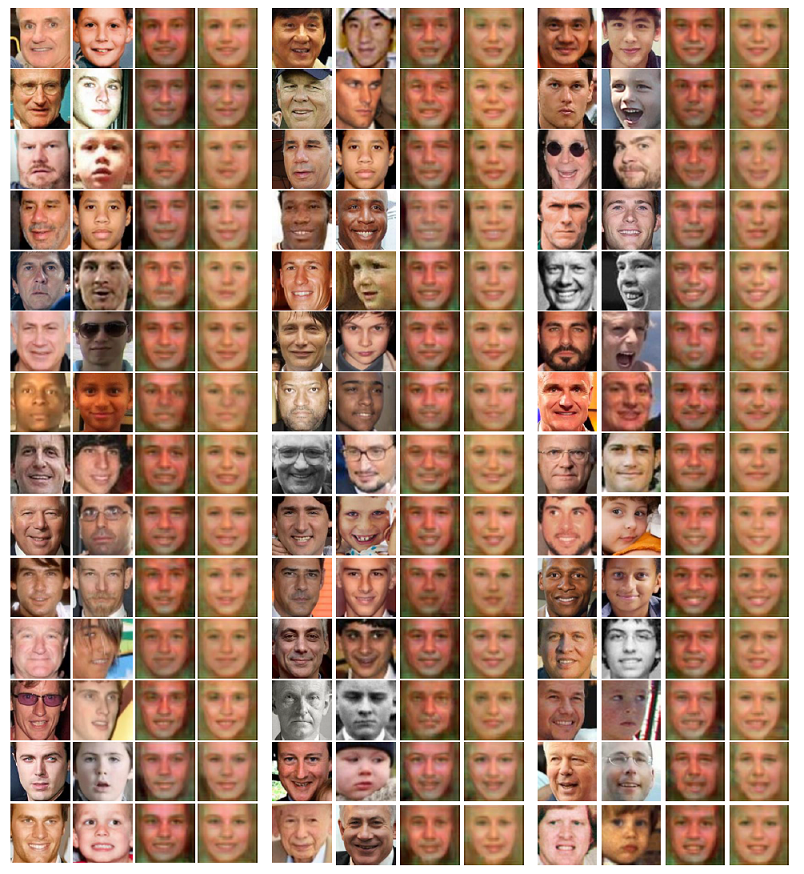}
\vspace{-0.3cm}
\caption{Additional results for father-son.}
\label{fig:fs_appendix}
\end{figure}

\begin{figure}[hbt]
\centering
\includegraphics[scale=0.75]{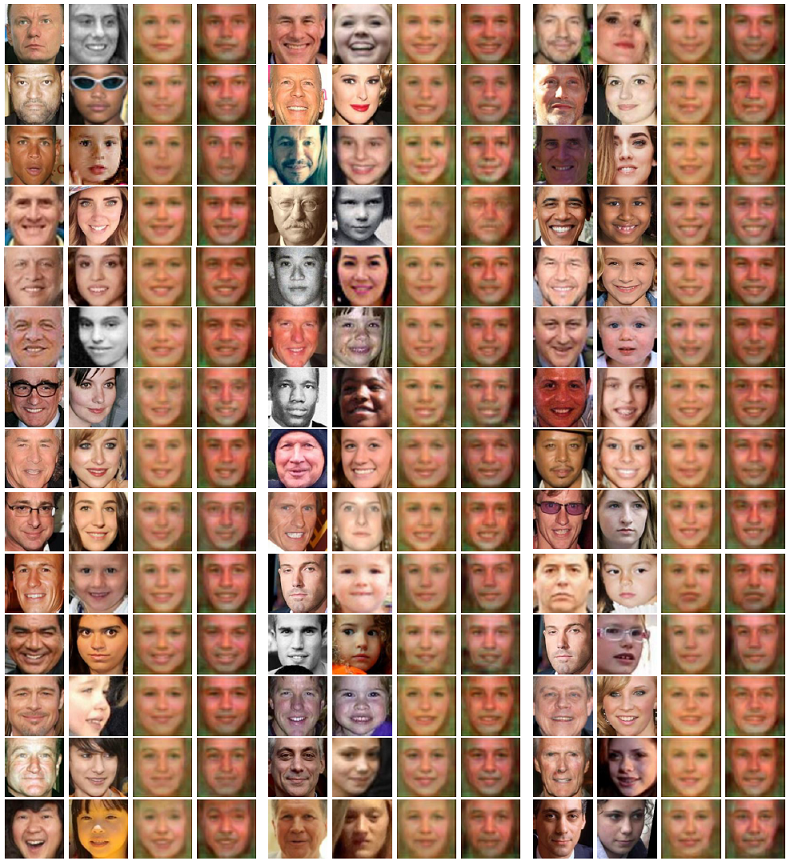}
\vspace{-0.3cm}
\caption{Additional results for father-daughter .}
\label{fig:fs_appendix}
\end{figure}

\begin{figure}[hbt]
\centering
\includegraphics[scale=0.75]{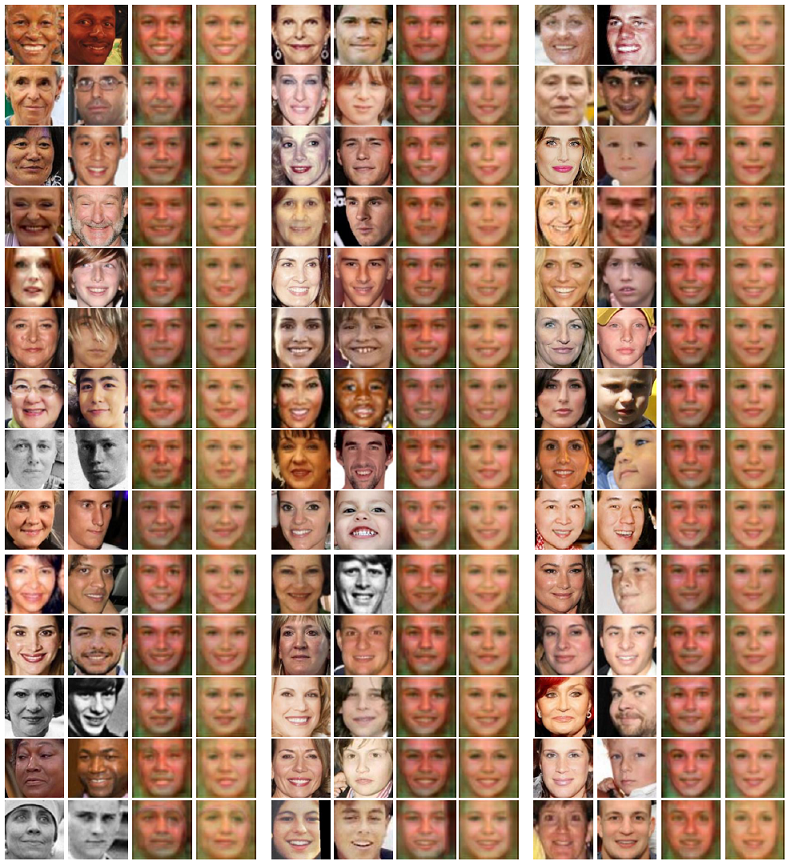}
\vspace{-0.3cm}
\caption{Additional results for mother-son.}
\label{fig:fs_appendix}
\end{figure}

\begin{figure}[hbt]
\centering
\includegraphics[scale=0.75]{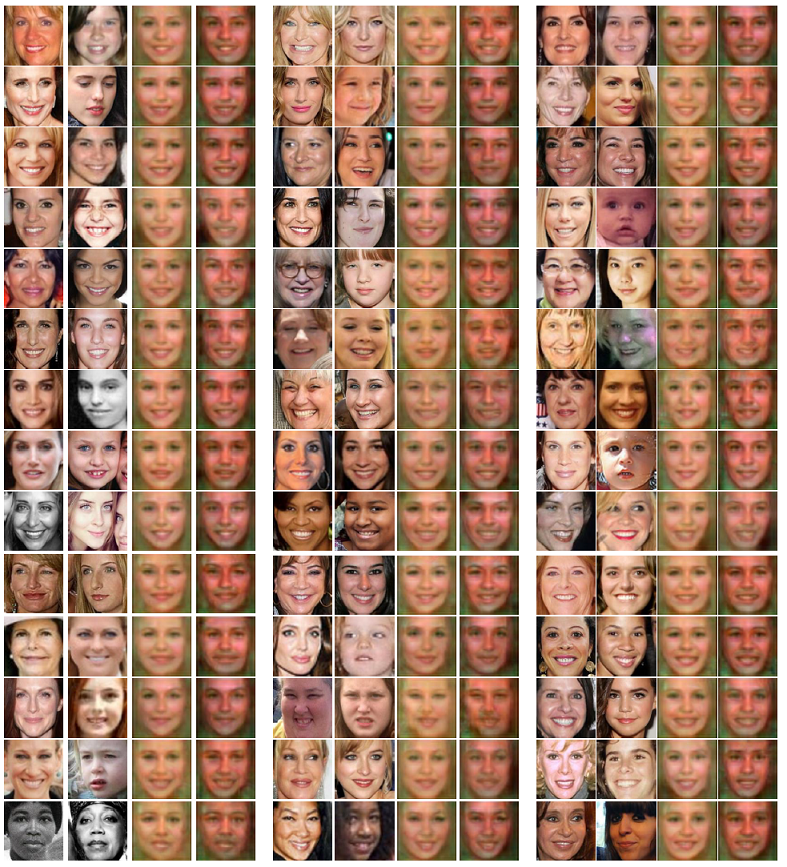}
\vspace{-0.3cm}
\caption{Additional results for mother-daughter.}
\label{fig:fs_appendix}
\end{figure}

\end{document}